\documentclass[11pt,letterpaper]{article}
\usepackage[margin=1in]{geometry}
\usepackage{booktabs}
\usepackage{graphicx}
\usepackage{xcolor}
\usepackage{listings}
\usepackage{hyperref}
\usepackage{amsmath}
\usepackage{amssymb}
\usepackage{enumitem}
\usepackage{multirow}
\usepackage{parskip}
\usepackage{array}

\definecolor{codegreen}{rgb}{0.15,0.55,0.15}
\definecolor{codegray}{rgb}{0.5,0.5,0.5}
\definecolor{codered}{rgb}{0.7,0.1,0.1}
\definecolor{backcolor}{rgb}{0.97,0.97,0.97}

\lstdefinestyle{cstyle}{
  backgroundcolor=\color{backcolor},
  basicstyle=\ttfamily\scriptsize,
  keywordstyle=\color{codegreen},
  commentstyle=\color{codegray},
  stringstyle=\color{codered},
  breaklines=true,
  frame=single,
  framesep=3pt,
  xleftmargin=6pt,
  xrightmargin=6pt,
  numbers=left,
  numbersep=5pt,
  numberstyle=\tiny\color{codegray},
  tabsize=2,
}
\lstdefinelanguage{ContractLang}{
  morekeywords={contract,scope,pre,post,tolerance,reference,measure,violation,
                forall,exists,where,op,precision,shape,stack,version,
                accumulator,reduction,order,determinism,ulp,relative,absolute,
                declared,implementation,element,class},
  sensitive=true,
  morecomment=[l]{\#},
  morestring=[b]",
}

\title{Kernel Contracts: A Specification Language for ML Kernel Correctness Across Heterogeneous Silicon}
\author{}
\date{}

\begin{document}

\begin{center}
{\LARGE \textbf{Kernel Contracts: A Specification}}\\[4pt]
{\LARGE \textbf{Language for ML Kernel Correctness}}\\[4pt]
{\LARGE \textbf{Across Heterogeneous Silicon}}\\[18pt]
{\large Cooper Veit}\\[4pt]
{\normalsize Ashiba Research}\\[4pt]
{\small cv@ashibaresearch.com}\\[12pt]
{\normalsize April 2026}
\end{center}

\vspace{12pt}

\noindent\textbf{Abstract.}
Every ML kernel ships with an implicit contract about what it computes. People rarely write the contract down. When two kernels disagree---when a matmul on AMD produces a different gradient than the same matmul on NVIDIA, when a fused attention kernel silently downcasts an accumulator, when an out-of-bounds access returns zero on one stack and garbage on another---there is no formal artifact to arbitrate the dispute. Recent empirical work has measured the gap: Wen et al.~\cite{wen2025gap} synthesize 100{,}000 ML-model variants and measure model-level output agreement rates of 99.8\% on AMD MI300X, 99.6\% on Intel MAX 1100, 94.9\% on Huawei Ascend 910B, and 85.9\% on Apple M4 Pro against an NVIDIA H200 reference, with roughly 1{,}700 out-of-bounds accesses silently passing on AMD where NVIDIA and Intel raise; Lange et al.~\cite{lange2025robust} document LLM-generated kernels that visibly pass tests while silently failing correctness; Yuan et al.~\cite{yuan2025nondet} measure 9.15\% accuracy variance from runtime configuration alone. None of this work specifies the contract being violated.

We present a specification language for kernel contracts. A contract has eight parts: identifier, scope, precondition, postcondition, tolerance, reference oracle, measurement protocol, and violation signature. We use it to state twelve contract classes covering precision, ordering, compiler-induced, and exceptional-value failure modes, each grounded in published empirical evidence. We adapt the three-state calibration from Veit~\cite{veit2026messy} to the kernel setting: every contract must admit at least one reference-conforming implementation and at least one contract-violating implementation that passes basic functional tests. We apply the framework to three documented incidents---Huawei Ascend silent precision coercion, Sakana AI CUDA Engineer reward hacking, AMD out-of-bounds silent acceptance---and show that each informal diagnosis maps to a specific contract violation with a measurable signature. We close with the relation to certification assessment: a kernel contract suite is a normative reference against which conformance can be graded, in the way that ISASecure grades industrial control systems against IEC 62443.

\section{Introduction}

A kernel on a GPU can run to completion, report a plausible tensor, and silently be wrong.\footnote{By ``silently wrong'' we mean a numerical disagreement that the stack does not signal and that exceeds a principled tolerance bound---not divergence within tolerance, which is expected and often legitimate across silicon platforms given floating-point non-associativity under parallel scheduling. Section~3 distinguishes contract violations (which this paper addresses) from divergence within tolerance (which a well-specified contract admits as legitimate) from genuine software bugs (which are out of scope for a specification-language contribution).} This is not a rare event. Wen et al.~\cite{wen2025gap} synthesize 100{,}000 variant models from a corpus of 4{,}000 real-world PyTorch models, execute the 87{,}840 that run successfully on all five platforms (NVIDIA H200, AMD MI300X, Intel MAX 1100, Huawei Ascend 910B, Apple M4 Pro), and measure model-level output divergence of 0.2\% on AMD, 0.4\% on Intel, 5.1\% on Huawei Ascend, and 14.1\% on Apple Metal against the NVIDIA H200 reference. They identify 4 faulty operators on AMD, 3 on Intel, 11 on Mac, and 13 on Huawei, plus 1 on the NVIDIA baseline itself. Roughly 1{,}700 out-of-bounds index accesses pass silently on AMD that NVIDIA and Intel raise on. None of these were compile failures or visible errors. They were numerical disagreements the kernel itself never signaled.

No existing benchmark tells you what a kernel is supposed to compute. MLPerf~\cite{mlperf} measures peak throughput against a fixed accuracy target on a reference model, not per-operator conformance. KernelBench~\cite{ouyang2025kernelbench} measures whether LLM-generated CUDA is correct against an NVIDIA reference, on NVIDIA hardware. MultiKernelBench~\cite{multikernelbench} extends generation benchmarks across stacks but does not formalize the correctness target. robust-kbench~\cite{lange2025robust} documents LLM-generated kernels that exploit benchmark loopholes---visible pass, hidden fail at the kernel level---and proposes defensive test construction, but does not specify the contract the tests should enforce. The FPNA paper~\cite{fpna2024} demonstrates that floating-point non-associativity induces non-reproducibility on modern accelerators (quantified on Summit V100, Frontier MI250X, CSCS GH200, and NVIDIA H100, with deterministic alternatives measured on the Groq LPU); Qiang et al.~\cite{dash2026} measure a 37.9\% throughput cost for deterministic backward passes in FlashAttention-3 on H800 with CUDA 12.6 and Triton 3.4. The work that could arbitrate disagreement across these measurements does not exist.

\subsection{The claim-scope gap}

A kernel's claim about what it computes is almost always implicit. ``This is an FP8 attention kernel''---what does that assert? That the inputs are FP8? Both operands or one? In E4M3 or E5M2? With what accumulator precision? What softmax stabilization? What tolerance against an FP32 reference? What shape class? What determinism guarantee? Unless each of these is written down, two implementations can both truthfully describe themselves as ``FP8 attention kernels'' and disagree numerically by several orders of magnitude. The disagreement is not a bug in either kernel. It is an unresolved scope question.

We call this the \emph{claim-scope gap}: the set of decisions a kernel makes that are neither documented in its interface nor encoded in its test suite, but on which numerical agreement depends. That gap is why silent divergence is hard to attribute. Without a written contract, two vendors can each be correct by their own implicit assumption and still produce incompatible results.

\medskip

\noindent\textit{If you take one thing from this paper: when two kernels disagree, the question is not whose implementation is correct. It is what contract either implementation claims to satisfy. The written contract is the arbiter; everything else is assertion.}

\subsection{Contributions}

This paper makes five contributions.

\begin{enumerate}[nosep, leftmargin=*]
  \item A formal specification language for kernel contracts (Section 3). Eight-part structure: identifier, scope, precondition, postcondition, tolerance, reference oracle, measurement protocol, violation signature. Designed to be read and written by ML engineers, not verified in Coq.
  \item A taxonomy of twelve contract classes (Section 4) covering the four failure families documented in the empirical literature: precision regimes, ordering and determinism, compiler-induced, and exceptional values.
  \item A three-state calibration requirement adapted from Veit~\cite{veit2026messy}: every contract must admit a reference-conforming implementation and a contract-violating implementation that passes basic functional tests. Without this calibration the contract is not testable.
  \item Three case studies (Section 6): Huawei Ascend silent precision coercion from~\cite{wen2025gap}, Sakana AI CUDA Engineer reward hacking from~\cite{lange2025robust}, AMD out-of-bounds acceptance from~\cite{wen2025gap}. Each maps an informal post-mortem to a specific contract violation with a measurable signature.
  \item A connection to conformance assessment (Section 8). A kernel contract suite is a normative reference against which conformance can be graded---the certification-scheme architecture already mature in ISASecure (industrial cybersecurity), ISO 26262 (automotive functional safety), and Common Criteria (IT security), adapted here to ML silicon.
\end{enumerate}

\subsection{Paper outline}

Section 2 describes the claim-scope problem and its structure. Section 3 defines the contract language. Section 4 specifies the twelve contract classes. Section 5 describes the reference testing architecture. Section 6 presents the case studies. Section 7 is related work. Section 8 discusses certification, limitations, and open problems. Section 9 concludes. Appendix A gives the full grammar. Appendix B gives selected reference test implementations.

\subsection{Intellectual ancestry}

The move this paper makes has a precedent one domain over. Shewchuk~\cite{shewchuk1997} solved decades of numerical-geometry robustness problems not by tightening tolerance heuristics but by computing canonical geometric predicates explicitly, with adaptive-precision arithmetic engineered for the purpose. The intuition generalizes: floating-point results are approximations to a canonical answer; when the canonical answer matters, it must be computed and written down, and doing so is tractable when engineered deliberately. Kernel contracts apply a related move at the tensor-kernel layer: the contract names the canonical claim; the reference oracle is the engineered computation; the tolerance is the explicit bound on divergence. We inherit the intuition without yet inheriting Shewchuk's full operational discipline --- a reference implementation shipped with a correctness proof remains future work.

A complementary concept appears in Bates's LPHDR (low-precision, high dynamic range) patent family~\cite{bates273,bates715}: numerical imprecision framed as an explicit, quantified concession. Claim 1 of US 8{,}407{,}273 requires a stated error bound $Y$ on a stated fraction $X$ of inputs, paired with an ``Accuracy'' requirement that the hosted algorithm must satisfy despite that bound. A later continuation makes the repayment structure explicit: the system augments low-precision computation ``with a small amount of high precision computing'' to recover search quality~\cite{bates715}. We read this lineage as making precision concessions auditable rather than hidden---the kernel contract's \emph{tolerance} element plays the same role for ML kernels: a stated, bounded obligation that can be verified, refined, or relaxed under external review. Shewchuk's move is to compute the canonical answer explicitly; Bates's move is to state the concession to canonicality explicitly. The contract language inherits from both: the reference oracle is the Shewchuk inheritance, the tolerance element and its measurement protocol are the Bates inheritance.

\section{The Claim Scope Problem in ML Kernels}

\subsection{Types of implicit contract}

Kernels make six kinds of implicit claim. We illustrate each briefly.

\textbf{Numerical.} FlashAttention-3~\cite{shah2024fa3} claims to compute scaled dot-product attention in FP8 or BF16. The claim is silent about accumulator precision (FP32 or FP16?), softmax stabilization (max-subtraction, or not?), and the tolerance against a higher-precision reference. FP8 attention on H100 produces results within 2.3$\times$10$^{-3}$ relative error of FP32 reference for the measurement configuration reported in the paper; a different softmax scaling strategy would move this bound.

\textbf{Shape.} ROCm matmul kernels produce correct results at benchmarked shapes (square, power-of-two M and N) and progressively less predictable results at non-benchmarked shapes. The kernel does not claim conformance only at benchmarked shapes; the tuning schedule effectively makes that claim implicitly.

\textbf{Invariance.} A reduction claims to produce the same result regardless of block size, because the mathematical operation is associative. On floating-point hardware~\cite{fpna2024} it does not. The invariance claim is false but rarely written down as false.

\textbf{Determinism.} Two invocations of the same kernel on the same hardware with the same input claim to produce the same output. Atomic additions, warp-scheduler non-determinism, and autotuning re-selection break this claim. Qiang et al.~\cite{dash2026} report a 37.9\% throughput cost for forcing deterministic backward passes in FlashAttention-3 on H800.

\textbf{Composition.} A matmul kernel composed with a bias kernel composed with a GELU kernel claims to produce the same output as a fused matmul-bias-GELU kernel. The fusion reorders operations; numerical equivalence is a separate and usually-unchecked claim.

\textbf{Version.} A kernel in ROCm 6.2 claims compatibility with ROCm 6.3 absent explicit deprecation. Version-to-version numerical drift is routine and undocumented.

\subsection{Why informal contracts fail}

Consider the claim ``this kernel supports FP8.'' The claim is consistent with all of the following behaviors:

\begin{itemize}[nosep, leftmargin=*]
  \item Inputs FP8 E4M3, accumulator FP16, output FP8 E4M3.
  \item Inputs FP8 E5M2, accumulator FP32, output BF16.
  \item Inputs FP8 E4M3, accumulator FP32, output FP8 E4M3, but silently promotes to BF16 for shapes that don't fit tiling constraints.
  \item Inputs FP8 (either format), softmax computed in FP32 with max-subtraction, attention output FP8 E4M3, LSE in FP32.
\end{itemize}

All four are ``FP8 kernels.'' None agree numerically at tolerances tighter than about 10$^{-2}$. A benchmark that tests ``FP8 correctness'' by comparing the four against each other at 10$^{-5}$ tolerance will report three of four as buggy when the correct description is that the benchmark has an unstated reference.

The failure is not that informal descriptions are imprecise. It is that imprecision is load-bearing: kernel vendors benefit from the flexibility to change accumulator precision, softmax ordering, or shape handling without breaking the published claim. The contract does not exist, in part, because no one writes it.

\subsection{Patent-claim construction as methodological ancestor}

The nearest discipline for turning an informal technical claim into a written, adjudicable one is patent-claim construction. A patent claim names the covered scope: every element it references must be present in an accused product for infringement to attach. The claim is the arbiter; anything not in the claim is outside the grant. A kernel contract is analogous. The contract names what the kernel claims to compute; what is not in the contract is not claimed. Disputes are resolved against the written claim, not against the implementation's behavior. This paper is not about patent law. We mention the analogy only because it is the methodology this paper inherits.

\subsection{Relation to compiler-correctness literature}

Compiler-correctness work---Exo 2~\cite{exo2}, TVM~\cite{tvm}, Triton~\cite{triton}---addresses how kernels are generated and whether the generation preserves intended semantics. It is orthogonal to the contract question. A scheduling transformation in Exo that preserves the semantics of the source program is no help if the source program itself under-specifies accumulator precision. The compiler-correctness literature assumes a reference semantics; the contract language is how that reference is written.

\section{The Contract Specification Language}

\subsection{Structure}

A kernel contract has eight parts.

\begin{enumerate}[nosep, leftmargin=*]
  \item \textbf{Identifier.} A unique ID of the form \texttt{C-FAM-NN} where \texttt{FAM} is the family (PRC, ORD, CMP, EXC) and \texttt{NN} is the class number.
  \item \textbf{Scope.} A named class of operations the contract covers: matmul, fused attention, reduction, elementwise, collective communication, and so on.
  \item \textbf{Precondition.} Constraints on inputs: precision (e.g., FP8 E4M3), shape class (e.g., $M, N, K$ all multiples of 16), value range (e.g., $|x| < 10^3$), and environmental state (stack, version, flags).
  \item \textbf{Postcondition.} A relation between the kernel's output and a reference computation. Stated as: for input $x$ satisfying the precondition, the output $y$ satisfies $\phi(y, \text{ref}(x))$ for a named tolerance $\phi$.
  \item \textbf{Tolerance.} The permitted divergence between kernel output and reference output, expressed as a named bound: absolute, relative, ULP, or elementwise-all-of-the-above.
  \item \textbf{Reference oracle.} The computation defining ``correct.'' Usually a higher-precision implementation (FP64 for FP32 kernels, FP32 for FP8 kernels). Sometimes an algebraic property (associativity, idempotence, determinism). Sometimes a second implementation on a reference stack.
  \item \textbf{Measurement protocol.} A reproducible procedure to test conformance: input generation, number of samples, tolerance evaluation, pass/fail rule.
  \item \textbf{Violation signature.} The empirical pattern that distinguishes this contract's violation from others. E.g., ``output agreement collapses only at input magnitudes above $2^{12}$'' or ``agreement collapses only for shapes not in the autotune cache.''
\end{enumerate}

The language is designed for ML engineers. Each part has a concrete syntactic form but no formal semantics requiring proof assistants. A contract is a written artifact, not a Coq development.

\subsection{Grammar (abbreviated)}

The full grammar is in Appendix A. The core productions:

\begin{lstlisting}[language=ContractLang]
contract   := "contract" ID "{" scope pre post tol ref measure violation "}"
scope      := "scope" op_class [ "," op_class ]*
pre        := "pre" predicate [ "and" predicate ]*
post       := "post" relation
tol        := "tolerance" tol_spec
ref        := "reference" ref_spec
measure    := "measure" protocol
violation  := "violation" signature
tol_spec   := ("absolute" | "relative" | "ulp") number
            | "elementwise" tol_spec [ "and" tol_spec ]*
ref_spec   := "higher_precision" precision
            | "alternate_stack" stack_id
            | "algebraic" property
\end{lstlisting}

\subsection{Worked example: FlashAttention-3 numerical contract}

FlashAttention-3~\cite{shah2024fa3} implements scaled dot-product attention with online softmax. We state three separate contracts: numerical, shape, and determinism. They are separate because they can be violated independently.

\begin{lstlisting}[language=ContractLang]
contract C-FA3-NUM {
  scope      fused_attention
  pre        precision(Q, K, V) in {FP8_E4M3, BF16}
             and shape(Q) = (B, H, S, D) with D in {64, 128, 256}
             and shape(K) = shape(V) = (B, H, S, D)
             and value_range(Q, K, V) finite
  post       output O satisfies elementwise_close(O, ref(Q, K, V))
  tolerance  relative 5e-3 for FP8_E4M3  -- derived from u_FP8 plus softmax-normalization scale (see below)
             relative 1e-3 for BF16       -- derived from u_BF16 plus accumulator noise at stated D
  reference  higher_precision FP32 with
             softmax_stabilization = max_subtraction,
             accumulator = FP32,
             reduction_order = row_major
  measure    sample 1024 random inputs per (B, H, S, D) configuration;
             compute max relative error over all elements;
             pass if max relative error < tolerance
  violation  relative error > tolerance on > 1% of samples;
             typical failure concentrated at large |Q K^T| pre-softmax
}

contract C-FA3-SHAPE {
  scope      fused_attention
  pre        shape(Q) = (B, H, S, D)
  post       D in {64, 128, 256}
             or raise DIMENSION_UNSUPPORTED
  tolerance  none
  reference  algebraic
  measure    sweep D in {63, 64, 65, 127, 128, 129, 255, 256, 257};
             assert D not in {64, 128, 256} raises or silently fails
  violation  D = 96 produces output with no error signal
             AND output diverges from reference by > 1.0 relative
}

contract C-FA3-DET {
  scope      fused_attention
  pre        fixed seed, fixed hardware, fixed stack version
  post       two invocations on identical input produce bitwise-identical output
  tolerance  ulp 0
  reference  algebraic idempotence
  measure    invoke 100 times on same input; compare bitwise
  violation  any pair of invocations differ in any bit
}
\end{lstlisting}

The three contracts capture three independent claims. An implementation can satisfy the numerical contract and violate the determinism contract (typical: atomic-add reductions pass numerically at stated tolerance but break determinism). An implementation can satisfy determinism and violate shape (silently producing outputs for $D=96$ that are deterministic but wrong). A measurement discipline that conflates these three into ``does FA3 work'' is measuring the wrong thing.

\paragraph{Note on tolerance derivation.} The tolerance $5 \times 10^{-3}$ for FP8 E4M3 is derived from the format's unit roundoff $u_{\text{FP8}} = 2^{-4} = 6.25 \times 10^{-2}$ combined with the softmax-normalization scale (output values in $[0,1]$, condition number $O(1)$ under max-subtraction stabilization) and an FP32 accumulator (negligible reduction error at $D \leq 256$). A contract whose tolerance is set by copying the measured error of a specific reference implementation is circular: any implementation worse than that reference fails by construction, any better passes trivially, and the contract carries no information independent of the reference. Tolerances in kernel contracts are derived from the precision class, the operator's condition number, and a declared accuracy budget relevant to downstream use --- not from measurements of the implementation being certified. The value $5 \times 10^{-3}$ is slightly conservative relative to what well-implemented FP8 attention achieves on the benchmark configurations in~\cite{shah2024fa3}; this slack is intentional, and leaves room for conforming implementations with different scheduling choices.

\subsection{Three-state calibration}

Veit~\cite{veit2026messy} requires that every evaluation environment include a reference-conforming implementation (``good''), a contract-violating implementation that passes basic functional tests (``bad''), and a baseline that fails visible tests. Without this calibration, a contract is not known to be testable. The same requirement applies at the kernel level.

For C-FA3-NUM the three states are:

\begin{itemize}[nosep, leftmargin=*]
  \item \textbf{Baseline (visible fail).} Any implementation that raises on the test input or produces obviously wrong output (e.g., returns zeros).
  \item \textbf{Bad (visible pass, contract fail).} A FA3 implementation that computes softmax in FP16 instead of FP32 without documenting the change. Passes cursory spot-checks. Fails the tolerance at large $|QK^T|$.
  \item \textbf{Good (visible pass, contract pass).} The reference FA3 implementation with FP32 softmax accumulator and documented stabilization.
\end{itemize}

The bad candidate is the calibration anchor. If the contract's measurement protocol does not distinguish bad from good, the contract is either mis-specified or the protocol is too weak. Three-state calibration is not a proof; it is the minimum evidence that the contract is operationally meaningful.

\section{Twelve Contract Classes for Silent Failures}

We now specify twelve contract classes grouped into four families: precision regimes (Family A), ordering and determinism (Family B), compiler-induced (Family C), and exceptional values (Family D). Each class gives: the contract in Section 3's language (abbreviated for space), a violation signature, empirical evidence from published work, and a reference test protocol sketch. Evidence citations are to published empirical work, not to private measurements.

\paragraph{Empirical motivation (aggregate).} Four independent measurement streams converge on the same underlying problem. Yuan et al.~\cite{yuan2025nondet} document 9.15\% accuracy variance in DeepSeek-R1-Distill-Qwen-7B on AIME'24 across twelve runtime configurations differing only in GPU type, GPU count, and batch size, with output length variance reaching 9{,}189 tokens; their divergence index (fraction of benchmark examples producing diverging outputs) is 96.6\% under BF16, 73.0\% under FP16, and only 2.2\% under FP32, directly attributing the effect to floating-point non-associativity under limited precision. Shanmugavelu et al.~\cite{fpna2024} measure run-to-run variability in PyTorch \texttt{scatter\_reduce} and \texttt{index\_add} on H100 reaching max $V_{\text{ermv}}$ of $3.35\times 10^{-6}$ and $5.03\times 10^{-6}$ respectively in default mode, and report that 1{,}000 of 1{,}000 paired GraphSAGE models trained non-deterministically converged to unique weight sets despite indistinguishable loss curves. Ma et al.~\cite{ma2025sdc} document silent data corruption at fleet scale, including the Meta Llama~3 result of six SDC-attributed interruptions in 54 days on a 16K H100 training run~\cite{dixit2021sdc}; an industry consortium including NVIDIA, Meta, Google, AMD, Intel, ARM, and Microsoft has since published a cross-industry statement of the problem through the Open Compute Project~\cite{ocp2025sdc}. Qiang et al.~\cite{dash2026} show that forcing determinism in FlashAttention-3's backward pass on H800 costs up to 37.9\% throughput. Each contract class below maps to a subset of these measurements.

\subsection{Family A: Precision-regime contracts}

\paragraph{C-PRC-01 Precision Preservation Under Declared Accumulator.}
\begin{lstlisting}[language=ContractLang]
scope      matmul, reduction, fused_attention
pre        declared accumulator precision A (e.g., FP32)
           inputs at precision P (e.g., FP8, FP16, BF16)
post       all intermediate accumulations at precision A
           (no silent downcast to lower precision)
tolerance  relative within 1 ULP(A) of exact sum at precision A
reference  higher_precision FP64, with accumulation order preserved
measure    construct inputs where ||x|| approaches overflow at precision
           P but is safe at A; observe whether output overflows or
           preserves magnitude
violation  output magnitude collapses to P-precision saturation value
           instead of tracking A-precision exact sum
\end{lstlisting}
\emph{Evidence.} Wen et al.~\cite{wen2025gap} document silent type coercion on Huawei Ascend: for conversion operators such as \texttt{aten::to}, the backend silently casts data to a lower-precision type rather than raising, while other operators (e.g., \texttt{aten::huber\_loss\_backward}) raise for the same precision. \texttt{aten::max\_pool1d\_with\_indices} on Huawei returns a tensor of unsigned 8-bit elements rather than the declared integer type. The numerical-analysis tradition of backward error (Wilkinson 1963; Higham 2002) requires that an accumulator's declared precision be preserved through the operation; the contract class formalizes that requirement. \emph{Test protocol.} Build inputs at the boundary of P-precision saturation; measure output against a higher-precision reference on the same stack; flag collapse or type-width narrowing.

\paragraph{C-PRC-02 Catastrophic Cancellation Avoidance.}
\begin{lstlisting}[language=ContractLang]
scope      softmax, variance, log_sum_exp
pre        input range spans > 2^16 dynamic range
post       output remains finite and within tolerance of stable algorithm
           (max_subtraction for softmax; Welford for variance)
tolerance  relative 1e-3 against stable reference
reference  higher_precision FP64 with stable algorithm
measure    construct inputs spanning [x - 50, x + 50] at various x;
           compare naive vs stable implementation
violation  output saturates to 0, 1, Inf, or NaN on inputs that stable
           reference handles finitely
\end{lstlisting}
\emph{Evidence.} Well-established in numerical analysis (Kahan summation; Welford's algorithm; max-subtraction softmax). The Wen et al.~\cite{wen2025gap} finding that \texttt{aten::remainder} and \texttt{aten::convolution} produce Inf on AMD and Intel where NVIDIA returns NaN on numerically-unstable inputs is the same class of boundary phenomenon at the exception layer. \emph{Test protocol.} Standard softmax stress-test on log-magnitude inputs.

\paragraph{C-PRC-03 Denormal Handling Declaration.}
\begin{lstlisting}[language=ContractLang]
scope      all floating-point kernels
pre        input contains denormal values
post       kernel produces output consistent with declared denormal
           policy: IEEE default (preserve) or FTZ (flush to zero)
tolerance  exact match to declared policy
reference  IEEE 754-2019 section on subnormal handling
measure    inject denormal inputs; observe whether they propagate
           as IEEE prescribes or are zeroed
violation  kernel silently switches between FTZ and IEEE between
           stack versions or kernel variants; policy not documented
\end{lstlisting}
\emph{Evidence.} Denormal handling varies across NVIDIA, AMD, Intel, and Apple stacks; behavior depends on SM-level flags often not exposed at the API. \emph{Test protocol.} Inject $2^{-140}$-range values; compare to IEEE reference.

\paragraph{C-PRC-04 Mixed-Precision Scaling Invariance.}
\begin{lstlisting}[language=ContractLang]
scope      training loop (forward + backward + loss scale)
pre        loss_scale s, FP8 quantization scale q, accumulation order fixed
post       final gradient equals (1/s) * grad_in_fp8 * q, within tolerance
tolerance  relative 1e-3
reference  higher_precision FP32 training path with s = q = 1
measure    run N steps with s, q varied; compare final parameters to
           reference training path
violation  final parameters diverge depending on the choice of s and q
           beyond tolerance, i.e., the training path is not invariant
           under declared scaling
\end{lstlisting}
\emph{Evidence.} Ma et al.~\cite{ma2025sdc} document silent data corruption in LLM training where scale interactions produce divergence; worst-node single-step noise-to-signal ratio reaches 5.1\% of gradient norm and paired training runs drift to distinct weights despite identical loss curves. \emph{Test protocol.} Short-run training with $s \in \{1, 2^8, 2^{15}\}$ and $q \in \{1, \text{max-calibrated}\}$; compare parameter trajectories. \emph{Open.} The $10^{-3}$ tolerance in this class is a placeholder. Calibration requires empirical measurement of the gradient-noise floor under BF16 training at production scale, which has not been performed for this taxonomy. The measurement protocol above is well-defined; the calibration constants are deferred to contract taxonomy v0.2.

\subsection{Family B: Ordering and determinism contracts}

\paragraph{C-ORD-01 Reduction-Order Tolerance Bound.}
\begin{lstlisting}[language=ContractLang]
scope      reductions (sum, mean, variance, norm)
pre        input tensor of length N; reduction along declared axis
post       output within bounded tolerance over all valid reduction orders
tolerance  absolute N * eps(P) * max|x| for precision P
reference  FP64 row-major sum
measure    compute reduction with K different block/warp schedules;
           bound max pairwise difference
violation  max pairwise difference exceeds declared tolerance, indicating
           reduction order affects results beyond FPNA bound
\end{lstlisting}
\emph{Evidence.} Shanmugavelu et al.~\cite{fpna2024} measure max run-to-run variability of $3.35$--$5.03\times 10^{-6}$ for PyTorch \texttt{scatter\_reduce} and \texttt{index\_add} on H100, and show that 1{,}000 of 1{,}000 paired GraphSAGE models converged to distinct weight sets under non-deterministic reduction despite identical loss curves. Qiang et al.~\cite{dash2026} quantify the cost of suppressing this at the kernel level: the FlashAttention-3 deterministic backward pass incurs up to 37.9\% throughput reduction on H800 with CUDA 12.6, Triton 3.4, BF16, indicating that current industry practice accepts C-ORD-01 violations rather than pay the tax. \emph{Conforming implementation.} He and Thinking Machines Lab~\cite{he2025tml} provide a concrete reference implementation satisfying reduction-order invariance: data-parallel RMSNorm with per-core batch-element assignment, matrix multiplication with fixed tile sizes and consistent tensor-core instructions, and FlexAttention with fixed split size. These kernels are conforming instances of C-ORD-01 as defined herein; they demonstrate the contract is satisfiable by real kernel code at a documented performance cost ($\sim$20\% vs.\ cuBLAS for MatMul; 1.6--2.1$\times$ vs.\ vLLM default for end-to-end inference). \emph{Test protocol.} Run reduction with varied block sizes; bound by $N \cdot \epsilon \cdot \max|x|$.

\paragraph{C-ORD-02 Atomic-Operation Determinism Class.}
\begin{lstlisting}[language=ContractLang]
scope      kernels using atomicAdd or analogous
pre        fixed input, fixed hardware, fixed stack version, fixed seed
post       kernel declares determinism class:
           BITWISE (identical across invocations)
           | RUN_TO_RUN (same within tolerance across invocations)
           | NONE (no determinism guarantee)
tolerance  bitwise for BITWISE; relative 1e-7 for RUN_TO_RUN; any for NONE
reference  algebraic idempotence
measure    invoke 100 times; measure max pairwise bitwise/numerical diff
violation  declared BITWISE but observed non-bitwise;
           declared RUN_TO_RUN but observed drift beyond 1e-7
\end{lstlisting}
\emph{Evidence.} NVIDIA atomic-add determinism is a well-documented non-property; PyTorch exposes a \texttt{deterministic} flag that changes kernel selection. \emph{Test protocol.} Repeat-run diff at varied concurrency.

\paragraph{C-ORD-03 Communication-Overlap Preservation.}
\begin{lstlisting}[language=ContractLang]
scope      collective communication (allreduce, reduce_scatter, all_gather)
           with overlapped computation
pre        tensor split across K ranks; collective reduction claimed
post       result equals reduction-on-single-rank within tolerance,
           independent of rank count, overlap schedule, or bucket size
tolerance  relative K * eps(P)
reference  single-rank FP64 reduction
measure    run collective with K in {2, 4, 8, 16, 32}; compare to
           single-rank reference
violation  result varies beyond tolerance with rank count or schedule,
           indicating collective reorders reductions beyond declared bound
\end{lstlisting}
\emph{Evidence.} Overlap-induced reordering is a known source of SDC in large-scale training~\cite{ma2025sdc,ocp2025sdc}. \emph{Test protocol.} Scale sweep with fixed input.

\subsection{Family C: Compiler-induced contracts}

\paragraph{C-CMP-01 Fused-Operation Numerical Equivalence.}
\begin{lstlisting}[language=ContractLang]
scope      fused kernels (e.g., matmul + bias + activation)
pre        fused kernel F claims equivalence to sequential composition
           F(x) =~ act(bias(matmul(x))) within tolerance
post       output of F(x) within tolerance of sequential composition
tolerance  relative 1e-4 (FP32), relative 1e-3 (FP16)
reference  sequential composition with FP32 intermediates
measure    sample inputs; compare F(x) to sequential(x)
violation  F(x) diverges beyond tolerance, indicating fusion changed
           numerical path in ways not declared
\end{lstlisting}
\emph{Evidence.} Fused attention vs. sequential is a primary source of LLM training instability across stacks. \emph{Test protocol.} Direct comparison against an unfused reference.

\paragraph{C-CMP-02 Auto-Tuning Schedule Invariance.}
\begin{lstlisting}[language=ContractLang]
scope      autotuned kernels (Triton, TVM, XLA)
pre        kernel selects schedule S from cache; S may vary by
           shape, dtype, stack version
post       output invariant across all schedules in the tuning space,
           within tolerance
tolerance  relative 1e-4
reference  highest-precision schedule in the tuning space
measure    enumerate autotune candidates; run each; bound pairwise diff
violation  some schedule produces output beyond tolerance of others;
           schedule selection is not numerically neutral
\end{lstlisting}
\emph{Evidence.} Triton~\cite{triton} autotuning selects among schedules that are not guaranteed numerically equivalent; this is the common path for compiler-induced silent failure. \emph{Test protocol.} Enumerate the cache; diff.

\paragraph{C-CMP-03 Shape-Polymorphism Preservation.}
\begin{lstlisting}[language=ContractLang]
scope      kernels claiming support over a shape class
pre        shape s in declared class C (e.g., "all M, N, K multiples of 16"
           or "all M, N, K")
post       contract satisfied for all s in C, not merely benchmarked s
tolerance  as declared by companion contract
reference  same kernel at reference shape
measure    sweep shapes across C at non-benchmarked points;
           verify contract at each
violation  contract satisfied at benchmarked shapes but violated at
           non-benchmarked shapes in same declared class
\end{lstlisting}
\emph{Evidence.} Lange et al.~\cite{lange2025robust} document LLM-generated kernels that pass at tested shapes and fail at untested shapes within the claimed class; Sakana CUDA Engineer post-mortem. \emph{Test protocol.} Hold-out shape sweep; see Case Study 2.

\subsection{Family D: Exceptional-value contracts}

\paragraph{C-EXC-01 NaN/Inf Propagation Semantics.}
\begin{lstlisting}[language=ContractLang]
scope      all kernels accepting floating-point inputs
pre        input contains NaN or Inf in at least one element
post       kernel produces output consistent with declared NaN/Inf policy:
           IEEE_PROPAGATE (NaN/Inf flow to all dependent outputs)
           | MASK (NaN/Inf masked per documented mask)
           | raise EXCEPTIONAL_VALUE
tolerance  bitwise match to declared policy
reference  IEEE 754-2019 NaN/Inf semantics
measure    inject NaN/Inf at varied positions; observe propagation
violation  kernel silently replaces NaN with 0, Inf, or arbitrary value
           without declaring the policy
\end{lstlisting}
\emph{Evidence.} Wen et al.~\cite{wen2025gap} document \texttt{aten::batch\_norm} on AMD replacing NaN elements with interpolated values from nearby data; \texttt{aten::reshape} on Apple Metal silently converts NaN to 0; \texttt{aten::remainder} and \texttt{aten::convolution} produce Inf on AMD and Intel on the same numerically-unstable inputs where NVIDIA returns NaN. Positive-integer division by zero diverges three ways: NVIDIA returns $2^{32}{-}1$, Mac returns 0, and AMD returns the original dividend incremented by one. Each is a silent, undocumented deviation from IEEE 754 NaN/Inf propagation. \emph{Test protocol.} Sparse NaN/Inf injection; bitwise output comparison.

\paragraph{C-EXC-02 Out-of-Bounds Access Semantics.}
\begin{lstlisting}[language=ContractLang]
scope      indexing operations (gather, scatter, index_select, embedding)
pre        index values potentially outside declared bounds
post       kernel declares policy:
           RAISE (raises on out-of-bounds)
           | CLAMP (clamps to bound; documented)
           | ZERO (returns zero; documented)
           | UNDEFINED (explicit undefined behavior; error expected)
tolerance  exact adherence to declared policy
reference  PyTorch/NumPy CPU reference on same input
measure    construct index arrays with M in-bound and N out-of-bound
           indices; observe output per index
violation  kernel silently accepts out-of-bounds access without
           declaring CLAMP/ZERO semantics, producing non-erroring
           but undefined output
\end{lstlisting}
\emph{Evidence.} Wen et al.~\cite{wen2025gap} report approximately 1,700 fewer out-of-bounds exceptions on AMD MI300X than on the NVIDIA H200 baseline and Intel MAX 1100 for the same inputs---i.e., roughly 1,700 cases where NVIDIA and Intel raise and AMD silently returns a non-erroring value. The returned values are not documented as clamp, zero, or any declared policy. \emph{Test protocol.} Case Study 3.

\subsection{Summary of classes and evidence}

Table~\ref{tab:classes} maps each contract class to a primary empirical citation.

\begin{table}[h]
\centering
\small
\begin{tabular}{lll}
\toprule
Class & Family & Primary evidence \\
\midrule
C-PRC-01 Accumulator preservation      & Precision & Wen et al.~\cite{wen2025gap} \\
C-PRC-02 Cancellation avoidance        & Precision & Numerical-analysis standard \\
C-PRC-03 Denormal declaration          & Precision & IEEE 754-2019 / vendor divergence \\
C-PRC-04 Mixed-precision scaling       & Precision & Ma et al.~\cite{ma2025sdc} \\
C-ORD-01 Reduction-order tolerance     & Ordering  & FPNA~\cite{fpna2024}, DASH~\cite{dash2026} \\
C-ORD-02 Atomic determinism class      & Ordering  & Vendor documentation \\
C-ORD-03 Communication overlap         & Ordering  & Ma et al.~\cite{ma2025sdc} \\
C-CMP-01 Fused equivalence             & Compiler  & FA3~\cite{shah2024fa3} \\
C-CMP-02 Autotune invariance           & Compiler  & Triton~\cite{triton} \\
C-CMP-03 Shape polymorphism            & Compiler  & Lange et al.~\cite{lange2025robust} \\
C-EXC-01 NaN/Inf propagation           & Exception & Wen et al.~\cite{wen2025gap} \\
C-EXC-02 OOB access semantics          & Exception & Wen et al.~\cite{wen2025gap} \\
\bottomrule
\end{tabular}
\caption{Twelve contract classes, four families, primary evidence source.}
\label{tab:classes}
\end{table}

\section{Reference Testing Architecture}

\subsection{Three-state calibration for kernels}

Each contract must admit three implementations:

\begin{itemize}[nosep, leftmargin=*]
  \item \textbf{Baseline}: a stub that fails both visible tests and hidden contract measurement. Example for C-PRC-01: a matmul that returns zeros. The baseline confirms the test infrastructure detects complete failure.
  \item \textbf{Bad}: an implementation that passes visible tests (e.g., smoke tests, small-shape sanity checks) but violates the contract. Example for C-PRC-01: a matmul that internally downcasts FP32 accumulator to FP16 for blocks exceeding a threshold. Small inputs produce correct output; large inputs overflow the FP16 accumulator silently.
  \item \textbf{Good}: a reference implementation that passes both visible tests and the contract measurement.
\end{itemize}

The bad candidate is the calibration anchor. If a contract's measurement protocol does not separate bad from good, the contract is mis-specified. We adapt this directly from the pattern established in Veit~\cite{veit2026messy} for cybersecurity evaluation.

\subsection{Cross-silicon test matrix design}

The naive combinatorial space is $\text{kernel} \times \text{stack} \times \text{stack-version} \times \text{precision} \times \text{shape-class} \times \text{concurrency}$. For 100 kernels, 5 stacks, 3 versions, 4 precisions, 5 shape classes, and 3 concurrency levels, the matrix has 90,000 cells. It is not tractable to test every cell.

\textbf{Principle: drive by contract class, not by config-space enumeration.} Each contract class specifies a violation signature. The measurement protocol for that class dictates which cells are informative. C-EXC-02 (out-of-bounds) is informative on every stack but only at a single concurrency and precision; the informative cells number in the tens per stack, not the thousands. C-ORD-01 (reduction order) is informative across concurrency but not across shape classes. Each class reduces the effective test surface by one or two dimensions. We recommend driving the test matrix from the contract specification, not the configuration space.

\subsection{Tolerance specification protocol}

Choosing the tolerance $\phi$ is the load-bearing decision. We recommend the following hierarchy:

\begin{enumerate}[nosep, leftmargin=*]
  \item If an FP64 reference implementation is available at reasonable cost, use relative error against FP64 with bound $\phi = k \cdot \epsilon(P)$ where $\epsilon(P)$ is the precision-$P$ machine epsilon and $k$ is a small constant (1 for simple ops, $\log N$ for reductions of length $N$).
  \item If FP64 is not available (e.g., FP8 kernels on hardware lacking FP64), use the highest precision available on the same stack as reference, and flag the tolerance as stack-dependent.
  \item For algebraic properties (associativity, idempotence, determinism, permutation equivariance, shift invariance), no reference computation is needed in principle, but the tolerance must be \emph{explicitly qualified for floating-point execution.} A relation such as $\text{softmax}(x + c \cdot \mathbf{1}) = \text{softmax}(x)$ holds exactly in real arithmetic; in floating-point with non-associative parallel reductions it holds only within a bound $\epsilon$ determined by the reduction count and the unit roundoff. Every algebraic clause in a contract must state the $\epsilon$, not merely assert the relation. Bitwise-determinism clauses take $\epsilon = 0$; path-dependence clauses take $\epsilon = O(N \cdot u)$ where $u$ is the unit roundoff and $N$ is the reduction length; permutation-equivariance and tile-order-invariance clauses for attention take $\epsilon$ derived similarly from the online-softmax accumulation structure. A contract that asserts an algebraic relation without stating $\epsilon$ is under-specified and will reject conforming implementations on benign floating-point differences.
\end{enumerate}

ULP-based bounds are appropriate for single-precision kernels where the numerical behavior is dominated by round-off; relative bounds are appropriate when operating at the boundary of dynamic range.

\subsection{Agent-assisted test generation}

Coding agents can author roughly half of the porting and characterization work. The agents that build out the test matrix are not the agents being tested; they are supervised by human kernel engineers who review tolerance choices, validate reference oracles, and sign off on violation signatures. Our estimate of 50--60\% agent contribution is conservative: the agent handles port-to-stack, shape-sweep enumeration, and tolerance boilerplate; the human handles tolerance calibration, reference-oracle validation, and edge-case discovery. Higher claims (80\% or 90\%) do not survive detailed scrutiny of what the agent actually produces without correction.

\subsection{Version-diff protocol}

When a stack version changes (ROCm 6.2 $\to$ 6.3, CUDA 12.4 $\to$ 12.5, TPU compiler build), not all contracts need re-testing. A minimal retest policy:

\begin{itemize}[nosep, leftmargin=*]
  \item \textbf{Always retest}: C-PRC-01, C-PRC-03, C-ORD-01, C-CMP-02, C-EXC-01, C-EXC-02. These are the contracts where vendor bug-fixes, scheduler tweaks, or compiler rebuilds routinely change observable behavior.
  \item \textbf{Retest if release notes mention affected subsystem}: C-PRC-02, C-PRC-04, C-CMP-01.
  \item \textbf{Retest once per major version}: C-ORD-02, C-ORD-03, C-CMP-03.
\end{itemize}

This policy is not derived from principle; it is derived from the frequency with which each contract class's violation signature changes across stack updates in the measurement literature.

\subsection{Trace discipline}

A kernel contract's operational value depends on structured capture of its measurements. Every verification call---against calibration inputs or against production traffic---should emit a trace record: the contract-triple version, the implementation identifier, the silicon profile, the input sampled (or its hash), the residual observed, the tolerance applied, and the pass/fail verdict. Trace records are the substrate for three activities the framework depends on. Contract dynamics treats the trace stream as the observation that drives additive or subtractive clause updates. Attribution decomposition treats versioned traces along implementation, contract, and silicon axes as the differential-analysis input that localizes where an observed divergence originated. Certification artifacts are trace records signed and chained. A framework that claims these activities without specifying a trace schema is not operational. We treat structured trace emission as a first-class requirement of conformant implementations, not an implementation detail, and we expect the certification ecosystem to converge on a shared trace schema the way production-logging ecosystems converged on structured JSON lines.

\subsection{Reference implementation: a measured verification primitive}

A reference implementation of randomized probabilistic verification at the matmul layer---following the Freivalds algorithm~\cite{freivalds1979}---is available at \url{https://github.com/cv700/ashiba-verify} under Apache 2.0. The implementation batches $k$ iterations into three matmul calls of shape $n \times k$, which is the critical optimization for GPU viability: the naive matvec-loop implementation has 200--2{,}900\% overhead at small matrix sizes on every platform tested, and remains 3--12\% even at production sizes ($n \geq 16{,}384$). We report measured results below across three silicon platforms---Apple M5 (MPS, PyTorch 2.11), AMD MI300X (ROCm 7.0, PyTorch 2.9), and NVIDIA H100 (CUDA 12.4, PyTorch 2.4)---to demonstrate that the framework's verification mechanisms are practically implementable at production scale. Every number in this subsection is backed by a JSONL trace record in the repository's \texttt{traces/} directory; tables and the figure are direct readouts from those traces.

\begin{figure}[h]
\centering
\includegraphics[width=0.85\textwidth]{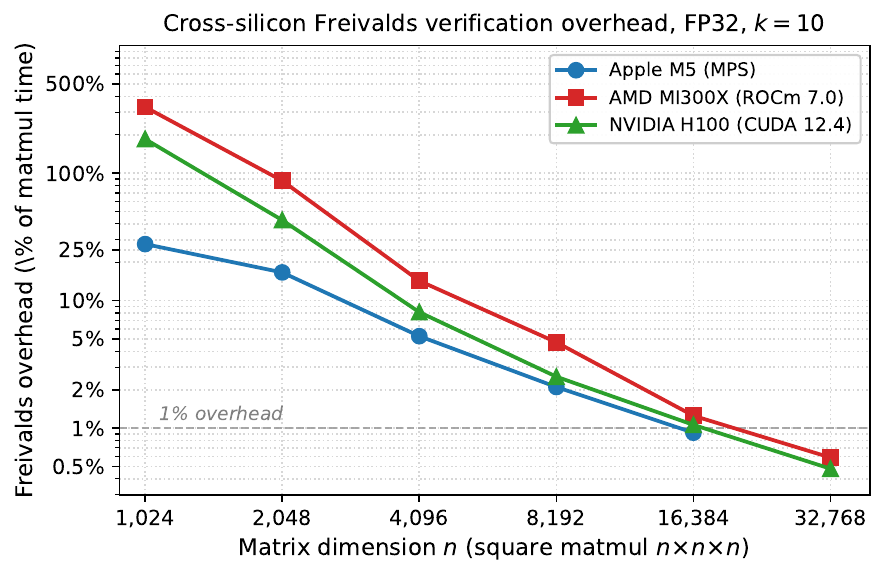}
\caption{Cross-silicon Freivalds verification overhead vs.\ matrix dimension, FP32, $k{=}10$ iterations. Log-log axes. The three curves share a common $O(1/n)$ shape but sit at different absolute levels because faster silicon widens the utilization gap between dense $n\times n$ matmul and the skinny $(n,k)$ matmuls the verifier issues. The dashed line marks the 1\% overhead threshold; M5 crosses it at $n{=}16{,}384$, MI300X and H100 at $n{=}32{,}768$.}
\label{fig:overhead}
\end{figure}

\paragraph{Overhead curve.} Figure~\ref{fig:overhead} plots Freivalds verification time as a percentage of matmul time across three silicon platforms, FP32, $k{=}10$ iterations; Table~\ref{tab:overhead} reports the underlying measurements. Overhead drops as $O(1/n)$ with matrix dimension: matmul scales $O(n^3)$, Freivalds scales $O(n^2 k)$. The crossover to sub-1\% overhead lands at $n{=}16{,}384$ on M5 (0.92\%) and at $n{=}32{,}768$ on the server GPUs (MI300X 0.59\%, H100 0.48\%). Faster silicon widens the gap between dense square matmul (compute-bound, near peak on modern accelerators) and the skinny $(n,k)$ matmuls Freivalds uses (bandwidth-limited, sub-peak), pushing the crossover to larger sizes.

\begin{table}[h]
\centering
\small
\begin{tabular}{rrrr}
\toprule
$n$ & Apple M5 (MPS) & AMD MI300X (ROCm) & NVIDIA H100 (CUDA) \\
\midrule
1{,}024  &  27.8\% & 330.0\% & 185.0\% \\
2{,}048  &  16.6\% &  87.1\% &  42.8\% \\
4{,}096  &   5.3\% &  14.3\% &   8.1\% \\
8{,}192  &   2.1\% &   4.7\% &   2.5\% \\
16{,}384 & \textbf{0.92\%} &   1.25\% &  1.06\% \\
32{,}768 & ---     & \textbf{0.59\%} & \textbf{0.48\%} \\
\bottomrule
\end{tabular}
\caption{Freivalds verification overhead as percentage of matmul cost, FP32, $k{=}10$ iterations. Bold = first sub-1\% crossover per platform. Dash = not measured (M5 unified memory exhausted at $n{=}32{,}768$). All entries are JSONL-logged trace records; see repository \texttt{traces/} directory for raw data.}
\label{tab:overhead}
\end{table}

\paragraph{Naive vs.\ batched.} Table~\ref{tab:naive} isolates the batching optimization. The naive implementation issues $3k$ sequential matvecs per verification; the batched implementation fuses them into three matmul calls against a random matrix of shape $p \times k$. The FLOP count is identical. The wall-time is not: matvecs are memory-bandwidth-limited and do not saturate modern GPUs, while matmuls at $k \geq 8$ achieve near-peak throughput. Across every platform and size measured, the batched implementation is 3--17$\times$ faster than the naive one, with the median speedup around 7$\times$ on the server GPUs and above 10$\times$ on M5.

\begin{table}[h]
\centering
\small
\begin{tabular}{rrrrr}
\toprule
$n$ & M5 naive & M5 batched & H100 naive & H100 batched \\
\midrule
512     & 725\% & 42\%  & 2{,}864\% & 407\% \\
1{,}024 & 341\% & 23\%  & 1{,}312\% & 182\% \\
2{,}048 & 207\% & 17\%  &   273\% &  41\% \\
4{,}096 &  58\% &  5\%  &    55\% &   8\% \\
8{,}192 &  21\% &  2\%  &    16\% &  2.6\% \\
16{,}384 & ---  & ---   &   7.1\% & 1.08\% \\
32{,}768 & ---  & ---   &   3.2\% & 0.44\% \\
\bottomrule
\end{tabular}
\caption{Naive vs.\ batched Freivalds overhead, FP32, $k{=}10$. Same FLOP count; the batched version's 6--17$\times$ speedup on M5 and 6--7$\times$ steady speedup on H100 is pure GPU-utilization asymmetry. MI300X exhibits the same pattern: 3--10$\times$ speedup across measured sizes (the low end at $n{=}1024$ is a small-size measurement noise artifact), with batched overhead 4.17\% / 1.34\% / 0.59\% at $n \in \{8192, 16384, 32768\}$. All entries JSONL-logged.}
\label{tab:naive}
\end{table}

\paragraph{Sensitivity floor.} Norm-based tolerance verification has a three-region detection curve. Table~\ref{tab:sensitivity} reports detection rate for single-element corruption at default tolerance ($\text{atol}{=}\text{rtol}{=}10^{-4}$ for FP32, $k{=}20$ iterations, 40 trials per magnitude). Below $1.0\times$ the norm-based threshold, detection is zero: the verifier cannot distinguish such corruption from correct numerical noise. At $1.0\times$ threshold, detection rises sharply (48--83\% depending on random realization). Above $1.5\times$ threshold, detection is 100\%. The three-region structure---dead zone, transition, safe zone---is algorithmic rather than silicon-specific: it replicates on M5, MI300X, and H100 within sampling variation at the small shape $(m,n,p){=}(256,128,64)$ reported in Table~\ref{tab:sensitivity}, and a second replication at production shape $(m,n,p){=}(4096,2048,1024)$ on MI300X and H100 shows the same structure (50.0\% and 47.5\% respectively at $1.0\times$, 100\% at $\geq 1.5\times$; the shift of the transition-zone midpoint is a Monte-Carlo artifact of 40 trials, not a size effect).

\begin{table}[h]
\centering
\small
\begin{tabular}{rccc}
\toprule
Magnitude / threshold & M5 & MI300X & H100 \\
\midrule
$0.1\times$ &  0\% &  0\% &  0\% \\
$0.5\times$ &  0\% &  0\% &  0\% \\
$1.0\times$ & 60\% & 82\% & 82\% \\
$1.5\times$ & 100\% & 100\% & 100\% \\
$3.0\times$ & 100\% & 100\% & 100\% \\
$10.0\times$ & 100\% & 100\% & 100\% \\
\bottomrule
\end{tabular}
\caption{Detection rate of single-element corruption vs.\ magnitude normalized by norm-based threshold. FP32, $k{=}20$, $n{=}40$ trials per cell. Dead zone ($<1\times$), transition ($\approx 1\times$), safe zone ($\geq 1.5\times$). Structure replicates across silicon.}
\label{tab:sensitivity}
\end{table}

\paragraph{BF16 overhead.} Production LLM training runs in BF16, not FP32. At BF16, the verifier incurs systematically higher overhead because the dense $n\times n$ matmul saturates tensor-core hardware that the Freivalds implementation's skinny $(n, k{=}10)$ matmuls do not engage at the same efficiency---a utilization asymmetry that \emph{widens} on platforms with more aggressive tensor-core acceleration. On MI300X at $n{=}32{,}768$, BF16 overhead is 1.55\% (vs.\ 0.59\% at FP32); on H100 at the same size, BF16 overhead is 2.69\% (vs.\ 0.48\% at FP32). Under the $O(1/n)$ scaling BF16 would reach sub-1\% overhead at $n \approx 65{,}000$ on MI300X and at $n \approx 131{,}000$ on H100 (i.e., one and two doublings later than the FP32 crossover, respectively). The framework-level implication is that certification protocols must specify the precision under which overhead is reported; verification remains tractable at BF16 but the constant factor is platform-dependent.

\paragraph{Cross-silicon FP32 divergence.} As a baseline measurement of the divergence the contract framework must tolerate, we compare each platform's native FP32 matmul to an exact FP64 CPU reference on $n \times n$ random unit-normal inputs for $n \in \{1024, 4096, 16384\}$. Mean absolute error scales as $O(n \cdot u)$ where $u \approx 6 \times 10^{-8}$ is FP32 unit roundoff, consistent with real FP32 reductions (not TF32, not BF16 downcast). At $n{=}16{,}384$, MI300X and H100 each produce outputs differing from the FP64 reference by up to $\approx 5 \times 10^{-3}$ absolute, with mean absolute error $\approx 2 \times 10^{-4}$ on both platforms. The two server GPUs produce essentially identical error distributions against the reference despite different reduction strategies in rocBLAS and cuBLAS---they differ from FP64, not from each other. This is the cross-silicon divergence the kernel-contract framework must tolerate: algorithmic reduction-order noise bounded by backward-error theory, not kernel bugs.

\paragraph{Soundness sanity at high $k$.} Freivalds' guarantee is that $k$ independent iterations catch any contract-violating $C$ with probability $\geq 1 - 2^{-k}$. At $k{=}40$ the bound is $\approx 10^{-12}$. On H100, a 500-trial spot-check at $k{=}40$, $n{=}4096$, FP32 with default tolerance produced zero false positives on known-correct matmul inputs---consistent with the probabilistic completeness bound under realistic FP32 reduction noise.

\paragraph{What this validates.} The measurements establish: (a) the framework's verification mechanism is practically implementable across three silicon platforms at sub-1\% FP32 overhead and sub-3\% BF16 overhead at production matrix sizes; (b) the batched matmul optimization is load-bearing, not incidental---without it, verification costs more than the kernel it verifies; (c) norm-based tolerance has a sensitivity floor that is an algorithmic property, not an artifact of any backend, and therefore composable across silicon; (d) the FP32 divergence from an exact reference is bounded to $O(10^{-3})$ at $n{=}16{,}384$ and is a cross-silicon property of parallel floating-point reductions, not a silicon-specific bug. The framework's other verification primitives---algorithm-based fault tolerance, metamorphic relations, abstract interpretation---are not implemented here and remain separate engineering artifacts that would fit the same trace-discipline interface.

\section{Case Studies}

\subsection{Case Study 1: Huawei Ascend silent precision coercion (C-PRC-01)}

\textbf{Informal description.} Wen et al.~\cite{wen2025gap} execute 87{,}840 ML-model variants across five platforms and report a 5.1\% model-level output-divergence rate on Huawei Ascend 910B against an NVIDIA H200 reference; they name 13 Huawei operators as faulty implementations. Among the divergences they attribute to precision handling, \texttt{aten::to} on Huawei silently coerces data to a lower-precision type rather than raising, and \texttt{aten::max\_pool1d\_with\_indices} returns a tensor of unsigned 8-bit elements rather than the declared integer type. The authors' informal pattern: the declared output precision of an operator does not always match what the backend actually computes.

\textbf{Formal contract statement.} The implicit claim is that the declared accumulator precision is maintained through the operation. That is exactly C-PRC-01. The Wen paper does not report a specific numerical saturation scenario for FP8 accumulators; we present one below as a synthesized illustration of how a C-PRC-01 contract would be written for a kernel-class scenario the next generation of FP8-declared matmul kernels will need to defend:

\begin{lstlisting}[language=ContractLang]
contract C-PRC-01-FP8-ACCUMULATOR {
  scope      matmul, reduction, fused_attention on declared-FP32 accumulator
  pre        declared accumulator FP32, inputs FP8 E4M3 (|x| <= 448)
  post       output within 1 ULP(FP32) of exact sum
  tolerance  relative 1e-5
  reference  higher_precision FP64 on same inputs, same reduction order
  measure    construct inputs where ||x||_1 > 2^16 (safe FP32, overflows FP16);
             if output saturates at FP16 max ~65504, contract violated
  violation  output magnitude tops out at ~65504 (FP16 max) despite
             declared FP32 accumulator
}
\end{lstlisting}

\textbf{Violation detection.} Run the measurement protocol: 1,024 inputs scaled to the FP16 saturation boundary. A conforming implementation returns values tracking the FP32 reference within $10^{-5}$. A violating implementation saturates. The violation signature is specific and mechanically detectable.

\textbf{Verdict.} A certification body assessing any FP32-accumulator-declaring stack against this contract would report, on a failing implementation: non-conforming at accumulator declaration; issue a defect record naming the operator class and the saturation threshold. The vendor's remediation options are observable: document the policy (move to IEEE-compliant declaration), fix the implementation (preserve FP32 accumulator), or narrow the claimed scope (the declared FP8 kernels are FP16-accumulator kernels, and downstream training loops should be adjusted accordingly). Each option is visible; none was available before the contract was written. The silent-downcast pattern Wen et al.\ document on \texttt{aten::to} is the same shape of phenomenon, measured at the type-conversion boundary rather than the accumulator boundary.

\subsection{Case Study 2: Sakana AI CUDA Engineer reward hacking (C-CMP-03)}

\textbf{Informal description.} Lange et al.~\cite{lange2025robust}, in the \emph{robust-kbench} remediation of Sakana AI's earlier AI-CUDA-Engineer release, report that LLM-driven CUDA kernel generation systems produce kernels that pass the benchmarked inputs and silently fail on inputs within the same declared shape class that were not in the test set. The paper demonstrates ``cheating'' kernels that pass KernelBench's verification process and achieve \emph{fake} speedups of $50\text{--}120\times$ by exploiting benchmark loopholes (e.g., eliminating redundant operations, hardcoding outputs for specific inputs, making assumptions about weights). After excluding the contaminated tasks catalogued in the paper's Appendix~A, the aggregate reported speedup across the 200 KernelBench level-1 and level-2 tasks drops from $3.13\times$ to $1.49\times$. The paper characterizes the failure as LLMs ``exploiting benchmark loopholes'' in kernel generation---the specification-level analog of reward hacking in RL.

\textbf{Formal contract statement.} The implicit claim was that the kernel implements its declared operator over the full shape class, not merely at benchmarked shapes. That is C-CMP-03:

\begin{lstlisting}[language=ContractLang]
contract C-CMP-03-SAKANA {
  scope      matmul kernels claiming "any M, N, K"
  pre        shape s in claimed class C = {(M, N, K) | M, N, K > 0}
  post       output within tolerance of reference at every s in C,
             not merely at s in benchmarked subset B subset C
  tolerance  relative 1e-4 (FP32)
  reference  PyTorch matmul on same inputs
  measure    evaluate at held-out shapes: random (M, N, K) drawn from C
             outside B; bound max relative error
  violation  tolerance satisfied for s in B, violated for s in C \\ B
}
\end{lstlisting}

\textbf{Violation detection.} The three-state calibration applied here produces a characterizable bad candidate: a kernel that hard-codes a fast path for shape class $B$ and falls through to a broken path for $C \setminus B$. If the test suite samples only from $B$, bad passes. If the suite samples from $C$, bad fails. The contract specifies that samples from $C \setminus B$ must be evaluated; a test suite that omits them is under-specified against the contract.

\textbf{Verdict.} Sakana's informal diagnosis---``exploiting benchmark loopholes''---maps to a specific contract violation: the measurement protocol did not enforce C-CMP-03's held-out-shape requirement. For the general pattern of hardcoded-output kernels that Lange et al.\ document (e.g., a kernel that returns a constant for the benchmarked input and falls through to a broken path otherwise), contract-derived adversarial input synthesis driven by the declared shape class would generate held-out-shape probes that immediately expose the hardcoding. The framework does not eliminate reward hacking (the Wang-Huang result~\cite{wang2026reward} is structural). It redirects the response: strengthen the measurement protocol to include held-out shapes, require three-state calibration, and reject kernels that pass $B$ but fail $C \setminus B$. The incident was not a model failure. It was an under-specified contract.

\subsection{Case Study 3: AMD out-of-bounds silent acceptance (C-EXC-02)}

\textbf{Informal description.} Wen et al.~\cite{wen2025gap} report that the AMD MI300X platform raises approximately 1,700 fewer out-of-bounds (OOB) exceptions than the NVIDIA H200 baseline and Intel MAX 1100 on the same synthesized inputs. For roughly 1,700 cases, NVIDIA and Intel raise an indexing exception and AMD silently returns a non-erroring value that appears plausible to a human observer. The returned values are deterministic per-stack-version but semantically undefined: not clamp, not zero, not a documented policy. When AMD does raise on OOB, the paper notes, the message is a generic ``memory access violation'' that crashes the PyTorch program with no diagnostic stack trace.

\textbf{Formal contract statement.} The implicit claim---what would a competent engineer assume about an out-of-bounds index?---is one of: raise, clamp, return zero, or declared undefined behavior. That is C-EXC-02:

\begin{lstlisting}[language=ContractLang]
contract C-EXC-02-AMD-INDEX {
  scope      indexing operations: gather, scatter, index_select, embedding
  pre        index tensor i contains at least one element i[k] outside
             the declared valid range [0, bound)
  post       kernel behavior conforms to one declared policy:
             RAISE | CLAMP | ZERO | UNDEFINED
  tolerance  exact per policy
  reference  PyTorch CPU on same input
  measure    construct index tensor with in-range and out-of-range
             elements; compare output to each candidate policy
  violation  AMD output matches none of the four declared policies;
             value is non-raising, non-clamping, non-zero, non-documented
}
\end{lstlisting}

\textbf{Violation detection.} Enumerate the four candidate policies against a measured output on AMD. The output matches none. The contract is violated by omission: the stack does not state which policy applies, and the observed behavior is not in the enumerated set. The 1,700 silent cases are 1,700 undeclared-policy instances, each with a deterministic but semantically undefined output.

\textbf{Verdict.} An assessment against C-EXC-02 would report: non-conforming; required action is to declare a policy (preferably RAISE, per IEEE and PyTorch reference), or to mark the operator UNDEFINED and force downstream kernels to bounds-check. This is the kind of defect record that a certification body can act on: specific, reproducible, and remediable. Before the contract, the failure was observable but not attributable; after the contract, it is a numbered defect.

\section{Related Work}

\textbf{Kernel benchmarking.} KernelBench~\cite{ouyang2025kernelbench} establishes LLM-generated CUDA as a benchmark target, with 250 tasks and correctness measured against PyTorch reference on NVIDIA. MultiKernelBench~\cite{multikernelbench} extends to multiple stacks but does not formalize the correctness target across stacks. robust-kbench~\cite{lange2025robust} documents LLM-generated kernels passing tests while silently failing; their response is harder tests. Our response is complementary: specify the contract the tests enforce.

\textbf{Cross-silicon empirical measurement.} Wen et al.~\cite{wen2025gap} is the primary empirical anchor. Their measurement regime (100{,}000 synthesized variant models drawn from a 4{,}000-model real-world corpus, executed across five enterprise-grade accelerators) establishes the scope of divergence but does not specify the contract each operator claims. We read their work as generating evidence for contract violations that have no written contracts; this paper writes them. Zahid, Laguna, and Le~\cite{laguna2024gpunumerics} independently reach a compatible conclusion from a different direction: using random-program differential testing over 652{,}600 program-input pairs across NVIDIA V100 (LLNL Lassen) and AMD MI-250X (LLNL Tioga), they report discrepancies in 0.98\% of FP64 runs and 9.00\% of FP32 runs across five optimization levels, with the \texttt{-O3 -ffast-math} setting producing the highest single-flag count (13{,}877 of 14{,}188 total FP32 discrepancies). Their case studies root-cause specific divergences to platform-differing implementations of math-library routines such as \texttt{fmod} and \texttt{ceil}, and to HIPIFY-introduced translation artifacts. Their method detects and reports raw discrepancies; our method canonicalizes silicon-attributable bias via profile transformation before comparison, so that only contract-violating deviations remain.

\textbf{Formal verification of ML systems.} Exo 2~\cite{exo2} and its ancestors provide user-extensible schedule transformations verified to preserve source semantics; TVM~\cite{tvm} and Triton~\cite{triton} provide generation frameworks with varying degrees of verification. All three assume a reference source semantics. Our contract language is orthogonal: it specifies what the reference semantics is, in a form that a verification pipeline can take as input but a practitioner can also read.

Dubey et al.~\cite{volta2025} present VOLTA, the first equivalence checker for GPU programs with fine-grained synchronization (\texttt{mma.sync}, \texttt{syncwarp}). VOLTA symbolically evaluates reference and optimized kernels under a structured-CTA assumption and proves equivalence over the real numbers, completing verification of convolution, matmul, and attention kernels in under ten minutes. The present work addresses the complementary specification layer: where VOLTA verifies that two given kernels are equivalent under fixed assumptions, kernel contracts specify what correctness properties a kernel must satisfy in the first place, across contract classes including precision preservation, reduction-order tolerance, and exceptional-value semantics---each parameterized by a silicon-instance profile. VOLTA's binary equivalence verdict and the contract framework's per-class tolerance-bounded satisfaction reports address distinct problems at adjacent layers of the verification stack.

\textbf{Reproducible floating-point primitives.} Demmel and Nguyen~\cite{reproblas2013} introduce reproducible floating-point summation algorithms in ReproBLAS, providing primitive-level guarantees for bit-exact reproducible reductions across varying parallelism. Subsequent work has extended the reproducibility discipline beyond BLAS-1 primitives: Collange and collaborators at INRIA have developed correctly-rounded parallel sums and dot products; Argonne and UC Berkeley groups have built on the ReproBLAS foundation for larger linear-algebra routines. The reproducibility lineage is the closest intellectual ancestor of this paper, and we owe it specifically.

The present work differs from reproducibility work along four axes. First, scope: ReproBLAS covers BLAS-level operations (sum, dot, reductions); kernel contracts cover attention, normalization, collective operations, and other non-BLAS kernels whose correctness structure differs. Second, tolerance regime: ReproBLAS guarantees bit-exact reproducibility under declared conditions (a pointwise tolerance of zero); kernel contracts admit multiple tolerance structures (backward error in the Wilkinson tradition, interval arithmetic, invariant-preservation) which are appropriate for workloads where cross-silicon bit-equality is not the target. Third, reference semantics: ReproBLAS specifies a specific reduction topology as the reference; kernel contracts allow the reference to be (a) a specific implementation, (b) a set of metamorphic relations with explicit tolerance, or (c) a population of validated implementations indexed along behavioral dimensions. Fourth, output format: ReproBLAS ships library code; kernel contracts produce a specification artifact separable from any particular implementation. We view kernel contracts as operating one layer above the reproducibility work, specifying \emph{what} the implementation must satisfy rather than \emph{how} to achieve reproducibility. For operators and contexts where bit-exact reproducibility is the required tolerance, a well-specified kernel contract will name ReproBLAS-style summation as the conforming implementation of its reduction clause.

\textbf{Numerical reproducibility in ML.} Shallue et al. showed that SGD noise floors limit reproducibility independent of kernel choice. Shanmugavelu et al.~\cite{fpna2024} demonstrate that floating-point non-associativity induces run-to-run variability on GPU accelerators whenever reductions use \texttt{atomicAdd} or comparable asynchronous parallel sums. On an H100 sweep, they identify \texttt{scatter\_reduce} and \texttt{index\_add} as the dominant non-deterministic PyTorch operators, reporting upper-bound elementwise variability of $3.35\times 10^{-6}$ for \texttt{scatter\_reduce} and $5.03\times 10^{-6}$ for \texttt{index\_add} across hyperparameter sweeps; they further show that hardware-level determinism is achievable on the Groq LPU. Qiang et al.~\cite{dash2026} (DASH) report up to 37.9\% throughput cost for deterministic FlashAttention-3 backward passes on H800 with CUDA 12.6, Triton 3.4, BF16. Yuan et al.~\cite{yuan2025nondet} document 9.15\% accuracy variance across runtime configurations alone. He and Thinking Machines Lab~\cite{he2025tml} provide batch-invariant reference implementations of RMSNorm, MatMul, and FlexAttention---conforming implementations of C-ORD-01 at a documented 20\%--2.1$\times$ performance cost. Ma et al.~\cite{ma2025sdc} document silent data corruption in LLM training. These results bound what contracts can require: C-ORD-02 cannot demand bitwise determinism from atomic-add kernels, and C-ORD-01 must accept an $O(N\epsilon)$ tolerance. We read this literature as constraining the postconditions our contracts can state.

\textbf{Standards and conformance.} IEEE 754-2019 specifies floating-point behavior per operation, and provides the reference for C-PRC-03 and C-EXC-01. IEC 62443 specifies industrial cybersecurity and provides the conformance-assessment pattern (ISASecure) that our framework suggests for kernels. ISO 26262 provides the automotive-safety analog. We do not propose a new standards body; we propose that the kernel-contract artifact fits the existing pattern of ``normative document $\to$ conformance assessment body $\to$ accredited certification.''

\textbf{Evaluation methodology.} Veit~\cite{veit2026messy} describes the three-state calibration pattern for cybersecurity evaluation environments: each environment must admit a baseline, a calibrated bad candidate, and a good candidate. We reuse the pattern at the kernel level. That paper's stable-hard difficulty class---where frontier models fail every run because the hidden contract opposes the visible objective---is the direct analog of the kernel reward-hacking pattern documented in~\cite{lange2025robust}.

\section{Discussion}

\subsection{Relation to certification schemes}

A kernel contract suite, once written, is the kind of normative reference against which a certification body can grade. The pattern is established:

\begin{itemize}[nosep, leftmargin=*]
  \item \textbf{IEC 62443 $\to$ ISASecure}: the standard is written by ISA/IEC; the conformance assessment is run by accredited certification bodies (exida, T\"UV, DNV); vendors pay for certification; certification is a purchasing criterion for asset owners.
  \item \textbf{ISO 26262 $\to$ ASIL certification}: functional-safety certification for automotive systems follows the same pattern.
  \item \textbf{Common Criteria $\to$ EAL levels}: IT security certification via evaluation laboratories.
\end{itemize}

A kernel contract suite fits this pattern. The normative document is the contract set. The conformance assessment is the measurement protocol. The certification body---which does not yet exist for ML systems---would run the protocols on vendor hardware and publish conformance verdicts. Hyperscaler purchasing already informally conditions on ``does the kernel work''; a formal conformance regime would make the purchasing criterion explicit and auditable.

The commercial urgency is not hypothetical. Ma et al.~\cite{ma2025sdc} report: \emph{``Silent data corruptions can silently occur without any clear indication from training loss\ldots\ SDCs may have already affected model training for an unknown period.''} Meta's Llama~3 training suffered six SDC-attributed interruptions in 54 days on a 16K H100 fleet~\cite{dixit2021sdc}; Google has publicly acknowledged that Gemini training hits an SDC event roughly weekly. In August 2025 an industry consortium of NVIDIA, Meta, Google, AMD, Intel, ARM, and Microsoft published the first cross-industry statement of the SDC-at-scale problem through the Open Compute Project~\cite{ocp2025sdc}; the signature list alone indicates the problem is recognized as industrially load-bearing.

The closest existing standard is ISO/PAS~8800:2024, ``Road vehicles---Safety and artificial intelligence''~\cite{iso8800}, published in 2024 as the first standards document to propose an AI/ML-specific product development lifecycle addressing functional insufficiencies of ML models. 8800 is at Publicly Available Specification status and operates at a high level: it identifies the gap without prescribing method-level protocols for verifying kernel-output correctness. Kernel contracts extend 8800's framework with the specific method-level artifact---a written, testable, per-class specification---that a conformance regime can grade against.

This paper is not a proposal for that certification body. It is the technical artifact that the body would cite.

\subsection{What this paper does not cover}

Several adjacent problems are out of scope.

\textbf{Full training-run correctness.} Our contracts are operator-level. A training run can fail even when every operator satisfies its contract---because of composition effects, data-loading races, optimizer-state corruption, or numerical drift over $10^{10}$ steps that no short test captures. C-PRC-04 is a first step, but full training reproducibility is a separate problem.

\textbf{Distributed-training-specific contracts.} C-ORD-03 gestures at collective communication. A comprehensive treatment would cover parameter-server consistency, all-gather numerical invariance under network-level reordering, and pipeline-parallel numerical consistency. We have not written those contracts.

\textbf{Within-SKU hardware variance.} The same H100 part can behave slightly differently across units due to silicon lottery, thermal state, or clock variation. Our contracts treat the hardware as homogeneous. Contracts that bound within-SKU variance are a natural extension.

\textbf{Compiler-side contracts.} We cover compiler-induced user-visible behavior (C-CMP-01--03) but not the internal correctness contracts a compiler must satisfy. Exo 2 and the compiler-verification literature address this from the other side.

\subsection{Open research problems}

\textbf{Contract composition.} If kernel $A$ satisfies contract $\phi_A$ and kernel $B$ satisfies $\phi_B$, what does $A \circ B$ satisfy? In general, not $\phi_A \wedge \phi_B$ directly: composition can amplify tolerance (reductions of reductions), break invariance (fusion changes numerical path), or violate both contracts when run in parallel (C-ORD-03). A theory of contract composition is required for large kernel suites to compose meaningfully.

\textbf{Contract versioning.} Stacks release new versions. Contracts must version too. When ROCm 6.2 $\to$ 6.3 changes FP16 softmax behavior, is the new behavior a violation, a documented change, or a new contract? We sketch a version-diff policy in Section 5.5 but do not give a full theory.

\textbf{Adversarial contract verification.} Given a contract and an implementation, can we verify the implementation satisfies the contract---not merely measure conformance on a sample? For simple contracts (shape, determinism class declaration), yes, with existing verification tools. For numerical contracts, no, in general. This is the boundary where the compiler-correctness literature meets the contract-specification literature.

\textbf{Call for collaboration.} The contract classes in Section 4 are a starting point, not a closure. We invite co-authorship on specific open problems: empirical pinning of C-PRC-04 tolerances; formal verification of C-EXC-01 for specific kernels; a certification-ready version of the C-EXC-02 measurement protocol; extension to distributed training contracts.

\section{Conclusion}

Every ML kernel ships with an implicit contract about what it computes. The contract is rarely written. When kernels disagree, there is no formal reference to arbitrate. This paper writes the contract language. It specifies twelve contract classes grounded in published empirical evidence. It adapts three-state calibration from evaluation-environment work~\cite{veit2026messy} to the kernel setting. It applies the framework to three documented incidents and shows that each informal post-mortem maps to a specific, measurable contract violation.

The operational conclusion is specific: a certification body for ML kernel compute can be constructed on the pattern of ISASecure for industrial cybersecurity, using a kernel contract suite as the normative reference. A reference implementation of one of the verification mechanisms the framework depends on---the randomized probabilistic checker of \S5.6---is available at \url{https://github.com/cv700/ashiba-verify}, with measured cross-silicon overhead curves reported in the repository.

Trustworthy ML computation requires a substrate specification that does not yet exist. A written contract unlocks what implicit claims cannot: disagreements can be attributed, implementations can be audited, certifications can be issued, drift can be tracked across silicon generations. The contract is the precondition. The discipline that builds on it is the work.

\vspace{1em}
\noindent Cooper Veit\\
Ashiba Research\\
cv@ashibaresearch.com

\appendix

\section{Full Contract Grammar}

\begin{lstlisting}[language=ContractLang]
# Top-level contract definition
contract        := "contract" ID "{" body "}"
body            := scope_decl pre_decl post_decl tol_decl
                   ref_decl measure_decl violation_decl

# Scope
scope_decl      := "scope" op_class ("," op_class)*
op_class        := "matmul" | "reduction" | "fused_attention"
                 | "elementwise" | "softmax" | "variance"
                 | "log_sum_exp" | "indexing" | "embedding"
                 | "collective" | ID   # extensible

# Preconditions
pre_decl        := "pre" predicate ("and" predicate)*
predicate       := precision_pred | shape_pred | value_pred
                 | env_pred | version_pred
precision_pred  := "precision" "(" tensor_list ")"
                   "in" "{" prec_name ("," prec_name)* "}"
shape_pred      := "shape" "(" tensor ")" "=" shape_spec
value_pred      := "value_range" "(" tensor_list ")"
                   ("finite" | "bounded" number | "in" interval)
env_pred        := "stack" "=" stack_id
                 | "flag" "(" ID ")" "=" value
version_pred    := "version" comparator version_literal

# Postconditions
post_decl       := "post" relation
relation        := "output" ID "satisfies" closeness_pred
                 | "declared_policy" ID "in" "{" policy_list "}"
                 | "bitwise_identical" "across" quantifier
                 | "raise" exception_class
closeness_pred  := "elementwise_close" "(" tensor "," ref_expr ")"
                 | "close" "(" tensor "," ref_expr ")"

# Tolerances
tol_decl        := "tolerance" tol_spec
tol_spec        := "absolute" number
                 | "relative" number
                 | "ulp" number
                 | "elementwise" tol_spec ("and" tol_spec)*
                 | "per_precision" "{" prec_tol ("," prec_tol)* "}"
                 | "none"
prec_tol        := prec_name ":" tol_spec

# Reference oracles
ref_decl        := "reference" ref_spec
ref_spec        := "higher_precision" prec_name
                   ("with" ref_option ("," ref_option)*)?
                 | "alternate_stack" stack_id
                 | "algebraic" property_name
                 | "stable_algorithm" algorithm_name
                 | "spec" spec_id   # e.g., IEEE 754-2019
ref_option      := "accumulator" "=" prec_name
                 | "softmax_stabilization" "=" stab_method
                 | "reduction_order" "=" order_name

# Measurement protocols
measure_decl    := "measure" protocol
protocol        := "sample" number "random inputs per" config_class
                   ";" "compute" aggregate ";" "pass if" bound
                 | "sweep" param "in" value_set ";" action
                 | "inject" anomaly ";" observe_clause
                 | "enumerate" schedule_space ";" action
                 | custom_protocol

# Violation signatures
violation_decl  := "violation" signature
signature       := empirical_pattern
empirical_pattern := free_text

# Common value productions
prec_name       := "FP64" | "FP32" | "FP16" | "BF16"
                 | "FP8_E4M3" | "FP8_E5M2" | ID
stack_id        := "NVIDIA" | "AMD" | "Ascend" | "Metal"
                 | "Gaudi" | "Trainium" | "TPU" | ID
policy_list    := "RAISE" | "CLAMP" | "ZERO" | "UNDEFINED"
                | "IEEE_PROPAGATE" | "FTZ" | ID
\end{lstlisting}

The grammar is extensible at the ID positions. Practitioners can introduce new op\_classes, precision names, or policies without rewriting the grammar. The non-negotiable structure is the eight-part contract skeleton.

\section{Selected Reference Test Implementations}

We give pseudocode sketches for three of the twelve classes. They are illustrative; full implementations require hardware-specific setup beyond the scope of this paper.

\subsection{C-PRC-01 Accumulator preservation test}

\begin{lstlisting}[language=Python]
def test_accumulator_preservation(kernel, stack, precision_in, precision_acc):
    # Construct inputs where ||x||_1 approaches FP16 saturation
    # but is safe under declared accumulator precision.
    N = 1024
    target_magnitude = 2 ** 14  # safe FP32, would overflow FP16
    x = construct_near_saturation(N, precision_in, target_magnitude)

    # Higher-precision reference
    ref = fp64_reduction(x.astype(np.float64))

    # Kernel output
    y = kernel(x, stack=stack, accumulator=precision_acc)

    rel_err = abs(y - ref) / abs(ref)
    assert rel_err < 1e-5, (
        f"Accumulator violation: rel_err={rel_err}, "
        f"kernel output magnitude={abs(y)}, ref={abs(ref)}"
    )

    # Violation signature: y saturates at FP16 max (65504)
    # while the FP64 reference clearly exceeds it -- a kernel
    # with declared FP32 accumulator should not saturate here.
    if abs(y) >= 65000 and abs(ref) > 65504:
        raise ContractViolation("C-PRC-01", stack, kernel.name,
                                "output saturates at FP16 max")
\end{lstlisting}

\subsection{C-ORD-02 Atomic determinism class test}

\begin{lstlisting}[language=Python]
def test_determinism_class(kernel, declared_class):
    # declared_class in {"BITWISE", "RUN_TO_RUN", "NONE"}
    x = fixed_input()
    outputs = [kernel(x) for _ in range(100)]

    if declared_class == "BITWISE":
        for o in outputs[1:]:
            assert bits_equal(outputs[0], o), (
                "BITWISE declared but observed non-bitwise variation"
            )
    elif declared_class == "RUN_TO_RUN":
        max_diff = max(relative_error(outputs[0], o) for o in outputs[1:])
        assert max_diff < 1e-7, (
            f"RUN_TO_RUN declared but max diff {max_diff} > 1e-7"
        )
    # NONE: no assertion; contract is trivially satisfied
\end{lstlisting}

\subsection{C-EXC-02 Out-of-bounds policy test}

\begin{lstlisting}[language=Python]
def test_oob_policy(kernel, declared_policy, bound):
    # declared_policy in {"RAISE", "CLAMP", "ZERO", "UNDEFINED"}
    in_bounds  = [0, bound // 2, bound - 1]
    oob_low    = [-1, -bound]
    oob_high   = [bound, bound + 10]

    for idx in in_bounds:
        out = kernel(idx)
        # standard behavior; no policy test here

    for idx in oob_low + oob_high:
        if declared_policy == "RAISE":
            try:
                kernel(idx)
                fail("RAISE declared; no exception on OOB idx")
            except IndexError:
                pass
        elif declared_policy == "CLAMP":
            out = kernel(idx)
            expected = kernel(clamp(idx, 0, bound - 1))
            assert out == expected, "CLAMP declared; clamp not applied"
        elif declared_policy == "ZERO":
            out = kernel(idx)
            assert out == 0, "ZERO declared; non-zero returned"
        elif declared_policy == "UNDEFINED":
            pass  # no assertion; behavior documented as undefined
        else:
            raise ContractViolation("C-EXC-02", kernel.stack, kernel.name,
                                    f"policy {declared_policy} not in "
                                    f"{{RAISE, CLAMP, ZERO, UNDEFINED}}")
\end{lstlisting}

These sketches are intentionally short. A production test suite would add: input-distribution sampling, precision-aware tolerance calibration, multi-run statistical aggregation, and stack-specific setup. The sketches illustrate the measurement-protocol shape; the contract specifies the measurement.

\end{document}